\documentclass[letterpaper, 10 pt, conference]{ieeeconf}  

\IEEEoverridecommandlockouts                              

\usepackage{graphicx} 
\usepackage{enumerate}
\usepackage{cite}
\usepackage{amsmath} 
\usepackage{amssymb}  
\usepackage{bm}

\usepackage{booktabs}

\usepackage{color}

\title{\LARGE \bf
Using Bayesian Optimization to Guide Probing of a Flexible Environment for Simultaneous Registration and Stiffness Mapping}

\author{ Elif Ayvali$^{1}$, Rangaprasad Arun Srivatsan$^{1}$, Long Wang$^{2}$, Rajarshi Roy$^{2}$, Nabil Simaan$^{2}$ and Howie Choset$^{1}$
\thanks{*This work was was supported by NRI Large grant IIS-1426655}
\thanks{$^{1}$E. Ayvali, R. A. Srivatsan, H. Choset are with the Robotics Institute at Carnegie Mellon University, Pittsburgh,PA 15213, USA
{\tt\small (eayvali@,rarunsrivatsan.,choset@cs.) cmu.edu}
       }
\thanks{$^{2}$L. Wang, R. Roy and N. Simaan are with the Dept. of Mechanical Eng., Vanderbilt University, Nashville, TN 37235, USA
        {\tt\small (long.wang, rajarshi.roy,nabil.simaan) @vanderbilt.edu}
				}
}

\DeclareMathOperator*{\argmin}{\arg\!\min}
\newcommand{\bs}[1]{\ensuremath{\boldsymbol{#1}}}
\newcommand{\norm}[1]{\left\lVert#1\right\rVert}
\usepackage[bordercolor=white,backgroundcolor=gray!30,linecolor=black,colorinlistoftodos]{todonotes}

\usepackage[normalem]{ulem} 

\newcommand{\realfield}{\hbox{I \kern -.35em R}}

\usepackage{subfigure} 

\begin{document}
\maketitle
\thispagestyle{empty}
\pagestyle{empty}

\begin{abstract}
One of the goals of computer-aided surgery is to match intraoperative data to preoperative images of the anatomy and add complementary information that can facilitate the task of surgical navigation.  In this context, mechanical palpation can reveal critical anatomical features such as arteries and cancerous lumps which are stiffer that the surrounding tissue. This work uses position and force measurements obtained during mechanical palpation for registration and stiffness mapping. Prior approaches, including our own, exhaustively palpated the entire organ to achieve this goal. To overcome the costly palpation of the entire organ, a Bayesian optimization framework is introduced to guide the end effector to palpate stiff regions while simultaneously updating the registration of the end effector to an \textit{a~priori} geometric model of the organ, hence enabling the fusion of intraoperative data into the \textit{a~priori} model obtained through imaging. This new framework uses Gaussian processes to model the stiffness distribution and Bayesian optimization to direct where to sample next for maximum information gain. The proposed method was evaluated with experimental data obtained using a Cartesian robot interacting with a silicone organ model and an \textit{ex~vivo} porcine liver.
\end{abstract}

\section{INTRODUCTION}

Surgeons performing minimally invasive surgery (MIS) offer their patients a shorter recovery time and reduced pain at a cost of increased technical difficulty. One of the drawbacks of MIS is the loss of direct sensory feedback. This loss can impede the detection and use of surface and stiffness features which can help the surgeon find correspondence between the intraoperative scene and the preoperative imaging data. Computer-aided surgery (CAS) was introduced to provide surgeons performing MIS  with functional and geometric information that can aid intraoperative navigation and execution of the preoperative plans. Mechanical exploration by palpation and tissue manipulation can provide complementary information about geometric constraints~\cite{bhattacharyya2013,boonvisut2014}, tissue characteristics~\cite{palacio2015} and variation in stiffness throughout the organ~\cite{Talasaz2013}, thus augmenting the information obtained through conventional image-based CAS. In this paper, we introduce tools for efficient information-guided exploration of an organ using mechanical palpation and integration of information, such as stiffness, to an \textit{a~priori} model through registration.\\

Mechanical palpation can facilitate the localization of arteries and other anatomically critical features which visually cannot be detected~\cite{Puangmali2008_review, Beccani2014, Faragasso2014}. 
Several groups previously proposed heuristic algorithms for autonomous exploration~\cite{Goldman2013} and segmentation of stiff features in compliant environments using classifiers ~\cite{Nichols2013}. These methods rely on the robot visiting all locations on a discretized grid on the organ's surface. Grid maps comes with the assumption of independence between the grid points ignoring the structural dependency in the environment. Accurate registration of the surgical tool to the coordinate frame of the preoperative model is one of the primary goals in incorporating intraoperative data, such as stiffness, into an \textit{a~priori} geometric model of the anatomy. Aforementioned previous works on palpation and exploration are not concerned with incorporating the stiffness information into the preoperative geometric data, a task that can help the surgeon in correlating preoperative models, like CT scans, with the current intraoperative scene.\\ 

To avoid the costly palpation of the entire organ and enable fusion of the stiffness information into an \textit{a~priori} geometric model of the anatomy, we introduce a method based on Bayesian optimization that minimizes the amount of probing required to reveal stiff features and registers the tool to the \textit{a~priori} model. The proposed algorithm is advantageous because it only visits regions that bring information gain, contrary to searching a discretized grid on the entire surface of the organ to find stiff features. \\

In the Bayesian formulation, a Gaussian process (GP)~\cite{rasmussen2006} is used to define a prior over the unknown stiffness distribution. Gaussian Processes provide a probabilistic description of the stiffness map and captures the variance of the stiffness distribution which helps guide the probing towards unexplored regions. The stiff regions correspond to regions near local maxima of the stiffness distribution and the Bayesian optimization finds the maxima of this unknown stiffness distribution by directing the probing to points that would result in maximum information gain in predicting the stiff regions. In a complementary effort, our collaborators at Johns Hopkins University are exploring different GP formulations to concurrently estimate surface geometry and stiffness for model reconstruction, using continuous palpation motion.  \\

We first introduce GP and Bayesian optimization in Section \ref{sec:Methods}, followed by the description of simultaneous registration and stiffness estimation. The experimental evaluations and results are given in Section~\ref{sec:Results}. 

\section{BACKGROUND}
\label{sec:Methods}
\subsection{Gaussian Processes}
A stochastic process is a collection of random variables, $\left\{Y : x \in \mathcal{X}\right\}$, indexed by elements from a set  $\mathcal{X}$, known as the index set. A Gaussian process is a stochastic process such that any finite subcollection of random variables has a multivariate Gaussian distribution \cite{rasmussen2004gaussian}. A GP , $f \sim \mathcal{GP}(\mu,k)$, is fully specified by its mean function $\mu : \mathcal{X} \rightarrow \mathbb{R}$ and a covariance function $k: \mathcal{X} \times \mathcal{X} \rightarrow \mathbb{R^+}$.

Intuitively, we can think of GP as a  distribution over functions.  Each random variable, $Y_i$, in a GP's collection is the distribution of function values at a point $x_i \in \mathcal{X}$.  Gaussian processes can be used for regression and to make predictions at a new point $x^* \in \mathcal{X}$ by defining a prior over functions. Given a set of $n$ observed inputs  $\bm{x} = [x_1, x_2, \dots, x_n]^T$ and corresponding outputs $\bm{Y} = [Y_1, Y_2, \dots, Y_n]^T$, the random variables $\bm{Y}$ are Gaussian distributed with mean $[\mu(x_1), \mu(x_2), \dots, \mu(x_n)]^T$ and covariance  matrix $\bm{K}$ whose elements are defined by a covariance function, $k(x_{i},x_{j})$ where $i$, $j \in [1,...n]$, that defines the covariance between $Y_{i}$ and $Y_{j}$. A commonly used covariance function is the squared exponential kernel defined as
\begin{equation}
k(x_{i},x_{j}) = \sigma_f \exp\left(\frac{-\left\|x_{i}-x_{j}\right\|}{2\ell^2}\right)
\end{equation}
where $\sigma_f$ is the variance of the process used as a scaling factor and $\ell$ is the length-scale of the kernel.

If we want to make predictions at a new set of $m$ points $\bm{x^*}$, the  joint distribution is given as
\begin{equation}
\begin{pmatrix} \bm{Y} \\\bm{Y}^* \end{pmatrix} \sim \mathcal{N}\Biggl(\begin{pmatrix} \bm{\mu}(\bm{x}) \bm{\mu}(\bm{x^*}) \end{pmatrix} ,\begin{pmatrix} \bm{K}(\bm{x},\bm{x}) & \bm{K}(\bm{x}^*,\bm{x})^T\\ \bm{K}(\bm{x}^*,\bm{x}) & \bm{K}(\bm{x}^*,\bm{x}^*) \end{pmatrix}\Biggr),
\end{equation}
In machine learning literature, $\bm{x}$ is called the training set and $\bm{x^*}$ is called the prediction or test set~\cite{rasmussen2006}. For simplicity, the prior mean function is generally assumed to be zero mean. The conditional (predictive) distribution $\bm{Y}^*$ can be computed using the conditioning rule of multivariate Gaussian distributions 
\begin{equation}
p(\bm{Y}_*|\bm{Y}) \sim \mathcal{N}(\bm{K}_*\bm{K}^{-1}Y,\bm{K}_{**}-\bm{K}_*\bm{K}^{-1}\bm{K}^T_*).
\end{equation}
where  $\bm{K}_*$ is the $m \times n$ training-prediction set covariance $\bm{K}(\bm{x}^*,\bm{x})$ and $\bm{K}_{**}$ is the $m \times m$ prediction set covariance matrix $\bm{K}(\bm{x}^*,\bm{x}^*)$.

We employ GP to model the stiffness distribution of the organ. The values of $\bm{Y}$ are the stiffness values associated with the probed points. The position of the points which are probed form the prediction set, $\bm{x}$, while $\bm{x}^*$ is the prediction grid which spans the surface of the organ. Note that GP is continuous and the prediction grid is only used to plot the stiffness distribution for visualization. 
By using GP, we assume the stiffness distribution changes smoothly across the organ and this smoothness is defined by the choice of the covariance function. In our formulation, we use a local deformation model for stiffness estimation. Therefore, a kernel with a fixed length-scale that is on the same order of the size of the palpation tool's tip is an effective choice.  We use the squared exponential kernel with $\sigma_f=1$ and $\ell=3$mm. The length-scale, $\ell$, determines how close two points have to be for the observation at those points to be correlated.

\setlength{\tabcolsep}{0.6mm}
\begin{table}[bh]
\caption{Notation}
\label{tb:Notation}
\centering
\begin{tabular}{cl}
\toprule
Symbol&Description\\ \midrule
$\bm{x}$& Coordinates of the probed points in the training set\\
$\bm{x}^*$& Coordinates of the grid points in the prediction set\\
$\bm{Y}$& The output at probed points $\bm{x}$ \\
$\bm{Y}^*$& The predicted output at $\bm{x}^*$ \\

$(\mu,\sigma)$&Mean and variance of the predictive distribution\\
$[\cdot]^R$& Entity expressed in tool's reference frame\\
$[\cdot]^C$& Entity expressed in CAD model's reference frame\\
$[\cdot]_0$& Initial value of the entity\\
$T$&Homogeneous transformation matrix\\
$\bs{n}$&Normal vector\\
$\bs{p}$&Coordinates of sensed position\\
$F$&Magnitude of sensed normal force\\
$c$&Stiffness at the probed point $x$\\
$\bs{\phi}$&CAD model with triangle faces and vertices \\\bottomrule
\end{tabular}
\end{table}

\subsection{Exploration and Exploitation with Bayesian Optimization}
Bayesian optimization is a powerful framework for global optimization of black-box functions~\cite{Jones1998}. It is most beneficial when the function does not have a closed-form expression and obtaining observations from the function is expensive. Bayesian optimization allows for prior beliefs about an unknown function to be updated via a posterior. Stiffness distribution of an organ can be thought of the unknown function we want to optimize whose maxima correspond to the stiff features.  In the Bayesian framework, we use GP to define a prior over the stiffness distribution. The sequential nature of the Bayesian optimization can help guide the sampling of the continuous search space. Sequential sampling requires selecting an acquisition function, also known as the utility function~\cite{Brochu2010}. Acquisition functions use the mean, $\mu(\bm{x})$, and variance, $\sigma(\bm{x})$, of the predictive distribution posterior to compute a function which shows the  most likely locations of the global maximum.
Acquisition functions such as probability of improvement~\cite{Jones2001}, expectation improvement (EI)~\cite{Jones1998}, and  upper-confidence based methods~\cite{Srinivas2010} have been developed to find the global maximum and to balance exploration with exploitation. EI provides global exploration of the search space and local exploitation. The EI acquisition function can be evaluated analytically and is given as~\cite{Jones1998}:

\footnotesize
\begin{equation}
EI(x)=
\begin{cases}
   (\mu(x)-Y^+) \Phi(Z)+\sigma(x) \phi (Z)     & \text{if }  \sigma(x)>0    \\
   0        & \text{if }  \sigma(x)=0
\end{cases}
\end{equation}
\normalsize
where
\footnotesize
\begin{equation}
Z=
  \begin{cases}
\frac{(\mu(x)-Y^+)}{\sigma(x)} & \text{if }  \sigma(x)>0    \\
   0        & \text{if }  \sigma(x)=0
\end{cases}
\end{equation}
\normalsize
$Y^+$ is the output at $x^+$ which is the current maximum of the sampled points. $\phi(.)$ and $\Phi(.)$ are the PDF and CDF of the standard normal distribution respectively.
   \begin{figure}[b!]
      \centering
      \includegraphics[width=0.8\columnwidth]{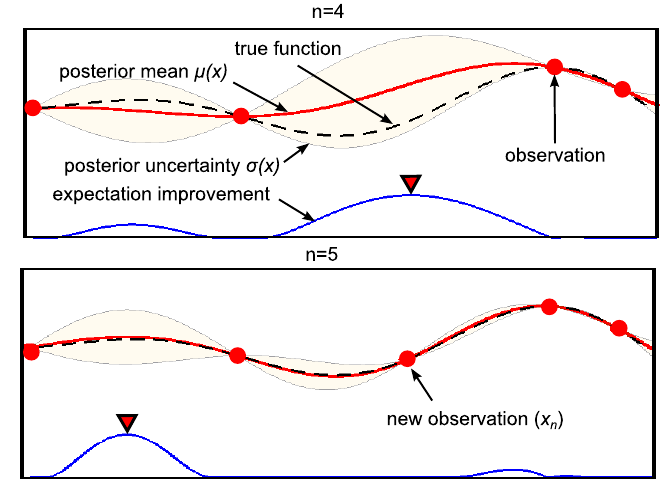}
      \caption{A 1-D example that starts with an initial training set of 4 points. Red triangle shows the maximum of the EI acquisition function which is the point that should be probed next.}
      \label{fig:EI}
   \end{figure}
	 \begin{figure}[b!]
      \centering
      \includegraphics[width=0.9\columnwidth]{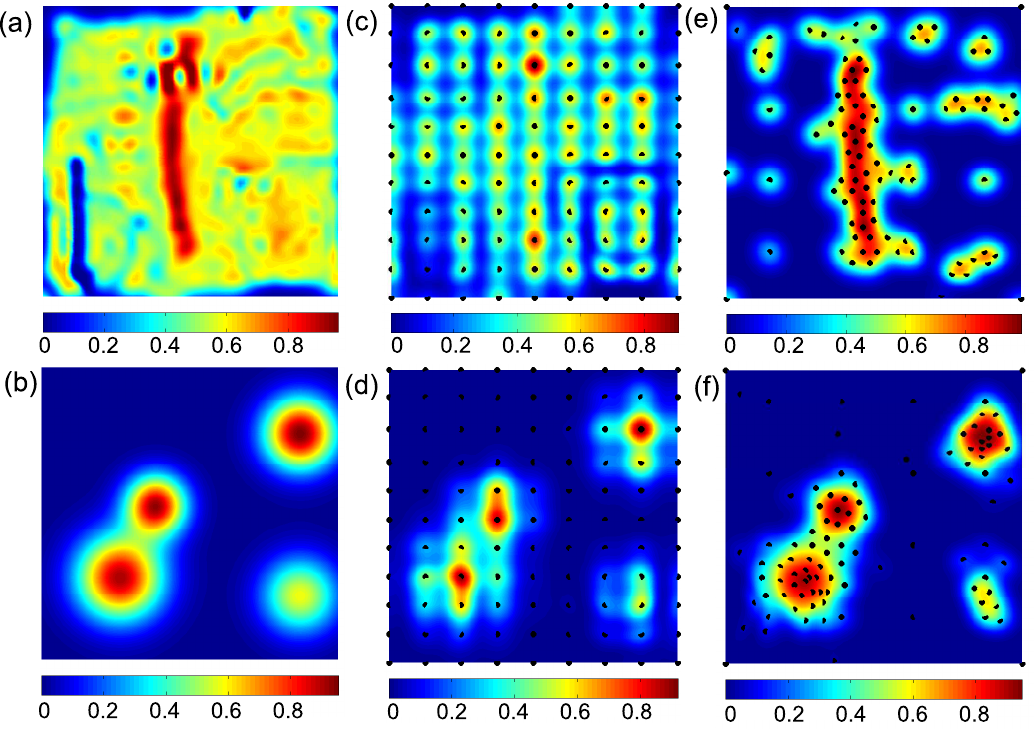} 
      \caption{Two examples that show the advantage of using EI as a sampling strategy: (a) and (b) show the ground truth stiffness map for a simulated data with an artery and simulated data with multiple stiff features, respectively; (c) and (d) show the maps obtained using uniformly sampling with 100 points; (e) and (f) show the corresponding maps obtained using EI with 100 samples.}
	\label{fig:GPRvsuniform}
   \end{figure}
	
Fig. \ref{fig:EI} illustrates an expected improvement scenario for a 1D example. Expectation improvement is generally used to find the global maximum of the unknown function and the search is terminated once a desired improvement is achieved. Note that after finding the global maximum, the algorithm still continues exploring other local maxima in the next iterations in an effort to reduce uncertainty and to find stiff features. In addition to using EI, we do pure exploration after every 5 samples and select a random point that has 90\% uncertainty. Such exploration is advantageous when there are multiple stiff features and the initial samples do not include points near stiff regions.  Preoperative information such as the size of the tumor or the width of the artery is useful to decide on the density of the samples in the initial set, however this is not the explored in this work. Interested readers can refer to ~\cite{Sobester2005} for discussions on the effect of the initial sample size in GP predictions.

Fig. \ref{fig:GPRvsuniform} demonstrates the stiffness map obtained using uniform sampling in comparison with the map sampled using EI for two different cases. With EI we can acquire useful information about the stiffness distribution compared to uniformly sampling the surface of the organ for the examples shown. 

\subsection{Registration and Stiffness Estimation}

Our group has previously developed a method for simultaneously estimating  the registration and stiffness distribution over the surface of a flexible environment using a Kalman filtering approach called  CARE~\cite{Siddharth2014} and a more recent model update method, called Complementary Model Update (CMU), that decouples stiffness estimation from registration, resulting in a more robust implementation. Similar to CARE, the CMU uses the force and position information obtained by interaction of the surgical tool with the organ to estimate the local stiffness and to register the organ to its preoperative model. It is assumed that the local surface deformations are only due to physical interaction of the surgical tool with the organ. We also assume that the surface of the organ is smooth and frictionless, thus the applied force is along the surface normal and increases with depth. Registration is performed by finding the transformation that takes the probed points defined in the tool frame to the corresponding points on the preoperative model of the anatomy. The preoperative model is a computer-aided design (CAD) model in the form of a triangular mesh.

The stiffness at a probed point is estimated using a best line fit between the relative sensed deformation depths and sensed forces:
\begin{equation}
\label{eq:linearstiffness}
c_i=L\left(\norm{(\bs{p}^R_{\beta})_i-(\bs{p}^R_{\gamma})_i},((F_{\beta})_i-(F_{\gamma})_i)\right)
\end{equation}
where $c_i$ is the stiffness of the $i^{th}$ probed point, $(\bs{p}^R_{\beta})_i$ and $(\bs{p}^R_{\beta})_i$ are the coordinates of two distinct sensed positions expressed in the tool reference frame, \textit{R}, corresponding to the $i^{th}$ probed point on the surface of the organ and $(F_\beta)_i, (F_\gamma)_i$ are the corresponding magnitude of sensed forces.
The registration estimate is obtained using the estimated stiffness, sensed position and magnitude of the sensed force:
\begin{equation}
\label{eq:objective}
\bs{T}=\argmin\limits_{\bs{T}} \sum \limits_{i=1}^n \norm{\bs{p}_{i}^C-\frac{\bs{n}_i^C (F_\beta)_i}{c_i}-\bs{T}(\bs{p}^R_{\beta})_i}
\end{equation} 
where $\bs{T} \in SE(3)$ is the homogeneous transformation that is to be estimated, $\bs{p}^C_i$ is the location of the probed point in the model's reference frame, given by \textit{C}, and the surface normal at the probed point is denoted by $\bs{n}^C_i$. More information about the CMU is provided in Appendix. 
\section{EXPERIMENTS AND RESULTS}
\label{sec:Results}

\subsection{Experimental setup}
\label{sec:Experiments}
\begin{figure}[htbp]
\centering
\subfigure[\label{fig:expsetup}]{\includegraphics[width =0.5\columnwidth]{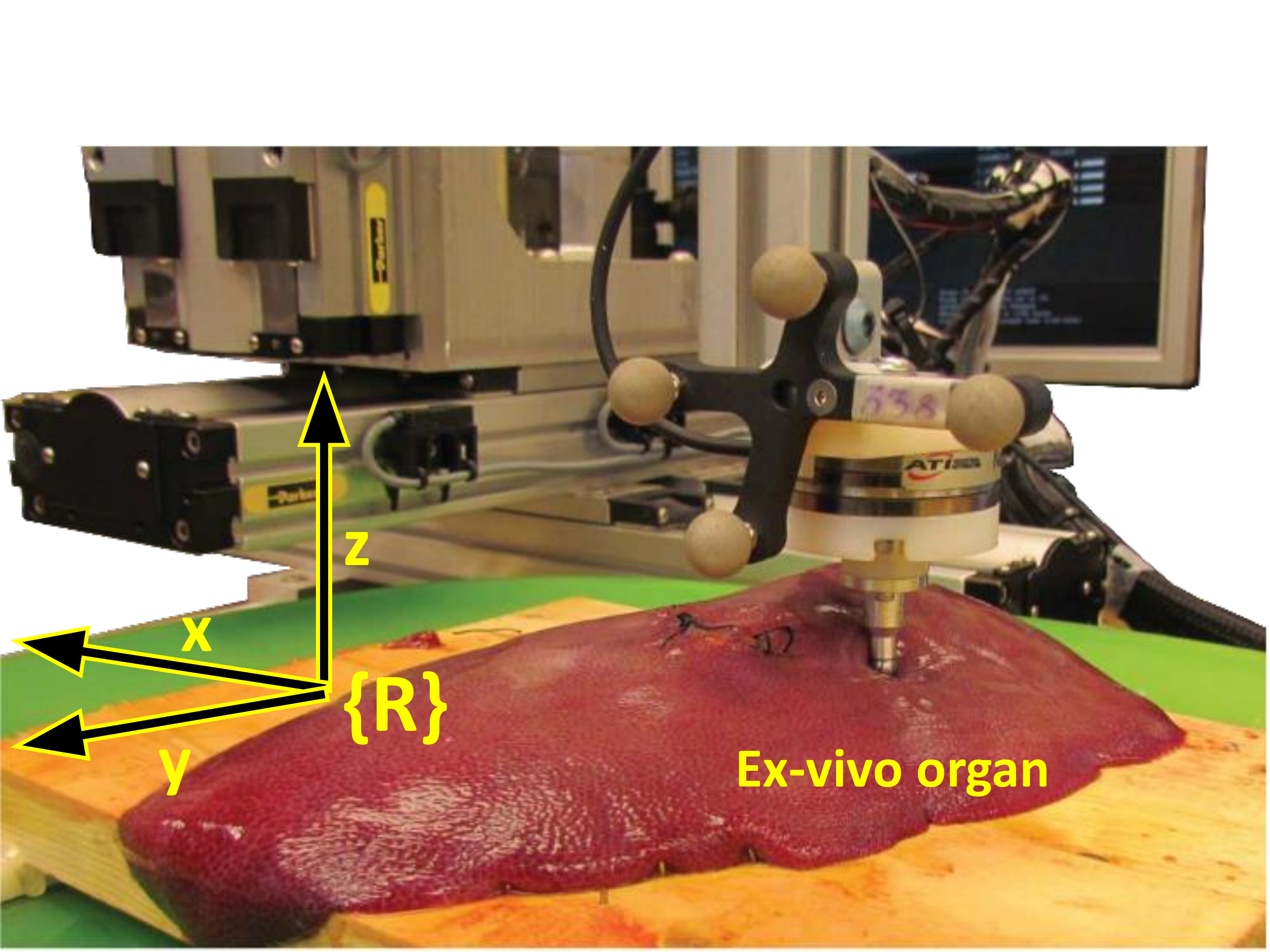}}\quad
\subfigure[\label{fig:ContactEst}]{\includegraphics[width =0.45\columnwidth]{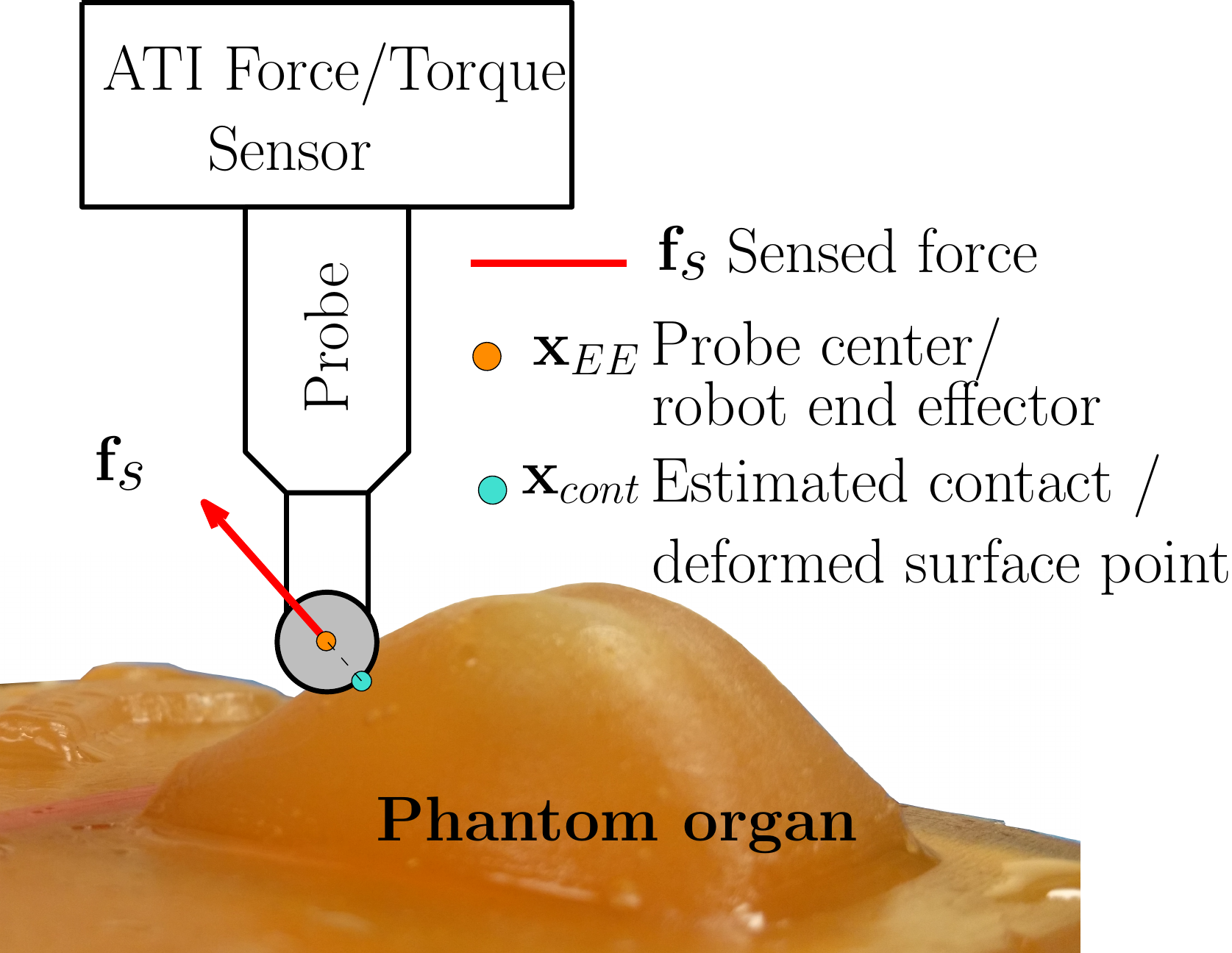}}
\caption{(a) Experimental setup used for \textit{ex~vivo} organ experiments. The tool frame, \textit{R}, is located at the end effector,(b) Contact location and surface norm estimation}
\end{figure}

A Cartesian robot with an open architecture controller was used to evaluate the framework proposed in this paper as shown in Fig. \ref{fig:expsetup}. The robot is equipped with an ATI Nano43 6-axis force sensor at the robot end-effector (EE) and is capable of executing the hybrid motion/force tasks described in Khatib \cite{khatib1987unified}. We assume that the robot end effector has been positioned above the organ. We do not assume any prior knowledge of the local surface normal.  

Experiments were conducted with a silicon phantom organ and an \textit{ex~vivo} porcine liver. In the silicon phantom experiment the top surface of the organ was lubricated and in the \textit{ex~vivo} experiment the organ surface was hydrated to reduce friction during probing. In both experiments, a target region was defined by the user. This region was then used as a reference to generate a uniformly distributed grid map with uniform spacing in the x-y plane of the tool's reference frame, \textit{R}. For a particular reference probing location, the following procedures were repeated automatically to obtain the force/deformation profile data.
\begin{enumerate}
    \item \textit{Making high force contact:}
    The robot controller is given a desired probing location $\mathbf{x}_\text{p}$ and a force magnitude of 0.5N.
    The hybrid force/motion controller decouples the combined commands into compatible (orthogonal) force/motion commands that direct the robot to achieve a desired position in the x-y plane and a stable contact force with the environment along the z axis.
    \item \textit{Estimating surface norm:}
    The contact location and surface normal estimation is shown in Fig. \ref{fig:ContactEst}.
    The surface normal, $\bs{n}$, is computed using the force sensed from the environment, $\bs{n}=\mathbf{f}_s/\|\mathbf{f}_s\|$, assuming the surface friction is negligible. 
    \item \textit{Finding low force surface contact point}:
    In this step, the robot first retreats swiftly away from the surface and then moves towards the surface, along the surface normal this time, to find the zero force intersection. An offset is applied from the robot EE to obtain the estimated contact point  as $\mathbf{x}_\text{cont}=\mathbf{x}_\text{EE} - \bs{n}r$ where $r$ is the radius of the robot end effector ball. A 9mm radius probe was used in the silicon examples and a 6.3mm radius probe was used in the \textit{ex~vivo} organ experiment.
    \item \textit{Probing and recording:}
    The robot is commanded in position control mode along the estimated surface normal direction with 0.3mm increments until 3mm probing depth. Hence, there are 10 position measurements, $\bm{p}$, and 10 force measurements, $F$, for each point we probe. 
\end{enumerate} 
\subsection{Bayesian Optimization Guided Probing}
\begin{figure}[htbp]
      \centering
      \includegraphics[width=\columnwidth]{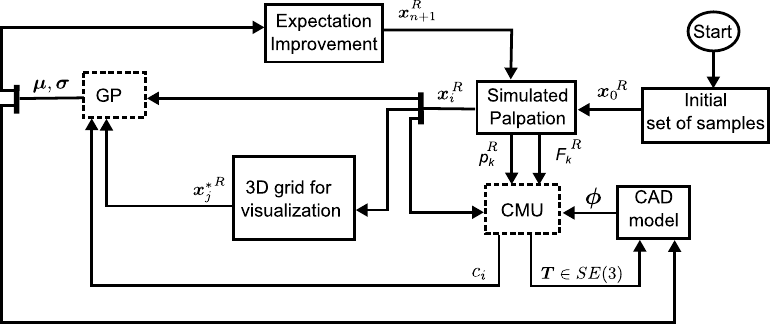} 
      \caption{Block diagram description of the Bayesian optimization guided probing}
	\label{fig:SCAR2flowchart}
   \end{figure}

To evaluate Bayesian optimization guided probing, we simulated experiments using the experimentally collected data. A block diagram description of the probing method is shown in Fig.~\ref{fig:SCAR2flowchart}.
Prediction of the stiffness distribution is carried out in the tool frame, \textit{R}. Initially, the actual registration is unknown, hence we do not know where the probed points correspond to in the preoperative CAD model. We assume palpation is carried out inside a region of interest (ROI). It is assumed that the initial set of samples, ${\bm{x}_0}^R$, include points on the boundary of the ROI as well as uniformly distributed points inside the ROI. The training set consists of previously probed points, ${\bm{x}_i}^R$, where $i=1,2,...,n$. We use gridfit function~\cite{gridfit} to  interpolate the previously probed points to form a dense grid inside the ROI, ${\bm{x}^{*}_{j}}^R$ where $j=1,2,...m$, to make predictions using GP. This grid is used to estimate the stiffness distribution for visualization. The stiffness value at a probed point, $c_i$, is estimated, in our case by the CMU, and corresponds to ${Y}_{i}^{R}$ at ${{x}_i}^R$ in the GP formulation. Based on the posterior of the predictive distribution given by $\mu$ and $\sigma$ the point at which the expectation improvement takes a maximum value is selected to be the next palpation point, ${x}_{n+1}^R$. As a new point is palpated, it is added to the training set and the prediction grid is regenerated. We use the updated registration estimate, $T$, to transform the probed points and their associated stiffness values to the corresponding points on the CAD model. This procedure enables displaying experimentally collected stiffness data on the preoperative CAD model.
\subsection{Evaluation}  
The proposed method was evaluated with various stiffness distributions and in the presence of measurement noise. We test the algorithm for four different scenarios:
\begin{enumerate} [1.]
\item Simulated data of an organ with a multimodal stiffness distribution.
\item Simulated data of an organ with a perturbed multimodal stiffness distribution and sensor noise.
\item Silicon phantom organ with an embedded mock artery.
\item \textit{Ex~vivo} porcine liver with a stiff inclusion.
\end{enumerate}

\begin{figure}[tbhp]
      \centering
      \includegraphics[width=0.75\columnwidth]{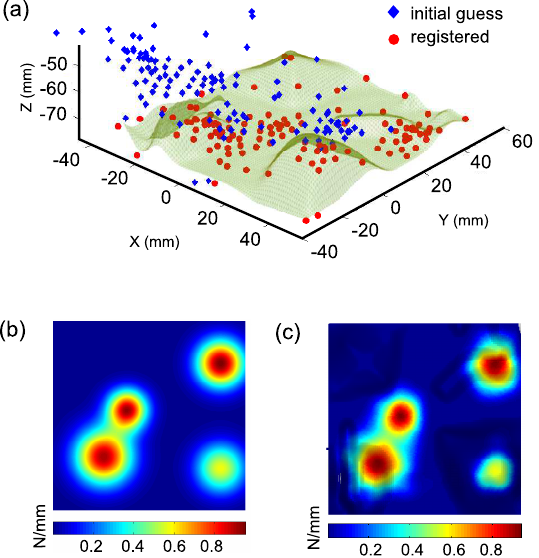}
      \caption{The algorithm starts with an initial set of of 19 points: 4 corners and 15 uniformly spaced points. The results are shown for 119 probed points. Example~1: (a)~Registration results, (b)~Ground truth stiffness map, (c)~Estimated stiffness map}
      \label{fig:GPRcases_part1}
   \end{figure}

The goal of Example~1 is to demonstrate that expectation improvement is effective in finding all the local maxima and not just the global maximum. We start with an initial set of 19 samples and terminate the palpation after 100 points. Fig.~\ref{fig:GPRcases_part1}(a) shows where we think the position of all the probed points are based on the initial registration guess and their registered position estimated by the CMU algorithm. The registered position of the probed points (deformed points) lie below the surface of the CAD model as expected; validating that the registration estimate is accurate. Fig.~\ref{fig:GPRcases_part1}(b) and (c) show the ground truth stiffness map and the predicted stiffness map, respectively. The predicted stiffness map captures the stiff features present in the ground truth stiffness map.

\begin{figure}[tbhp]
      \centering
      \includegraphics[width=0.8\columnwidth]{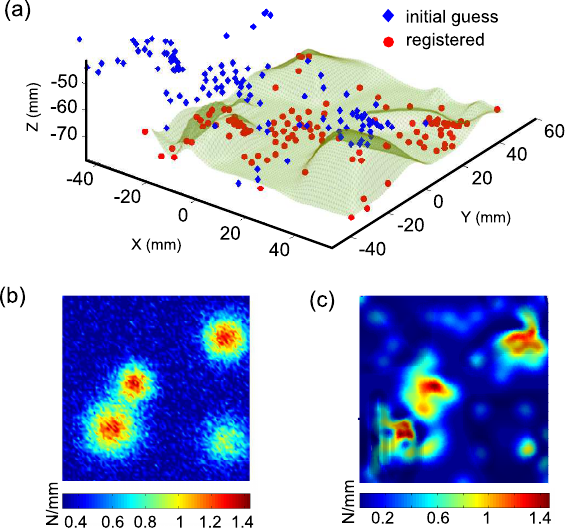}
      \caption{Example~2: (a)~Registration results, (b)~Ground truth stiffness map, (c)~Estimated stiffness map. The algorithm starts with an initial set of 19 points: 4 corners and 15 uniformly spaced points. The results are shown for 119 probed points.}
      \label{fig:GPRcases_part2}
   \end{figure}

Example~2 shows the effect of noisy sensor measurements in the prediction of the stiffness distribution. The ground truth stiffness map of Example~2 was obtained by perturbing the stiffness distribution for Example~1. An artificial sensor noise with $(\mu_{\bm{x}},\sigma_\textbf{x})$ =~(0, 0.3mm) was added to the sensed position  and $(\mu_F,\sigma_F)$ =~(0, 0.1N) was added to the sensed force to simulate a more realistic scenario. Fig.~\ref{fig:GPRcases_part2}(a) shows the registration results. Fig.~\ref{fig:GPRcases_part2}(b) and (c) show the ground truth stiffness map and the predicted stiffness map, respectively. In the presence of noisy sensor measurements, the algorithm still reveals the stiff features and the registration parameters converge to the correct values.
		 
\begin{figure}[tbhp]
      \centering
      \includegraphics[width=\columnwidth]{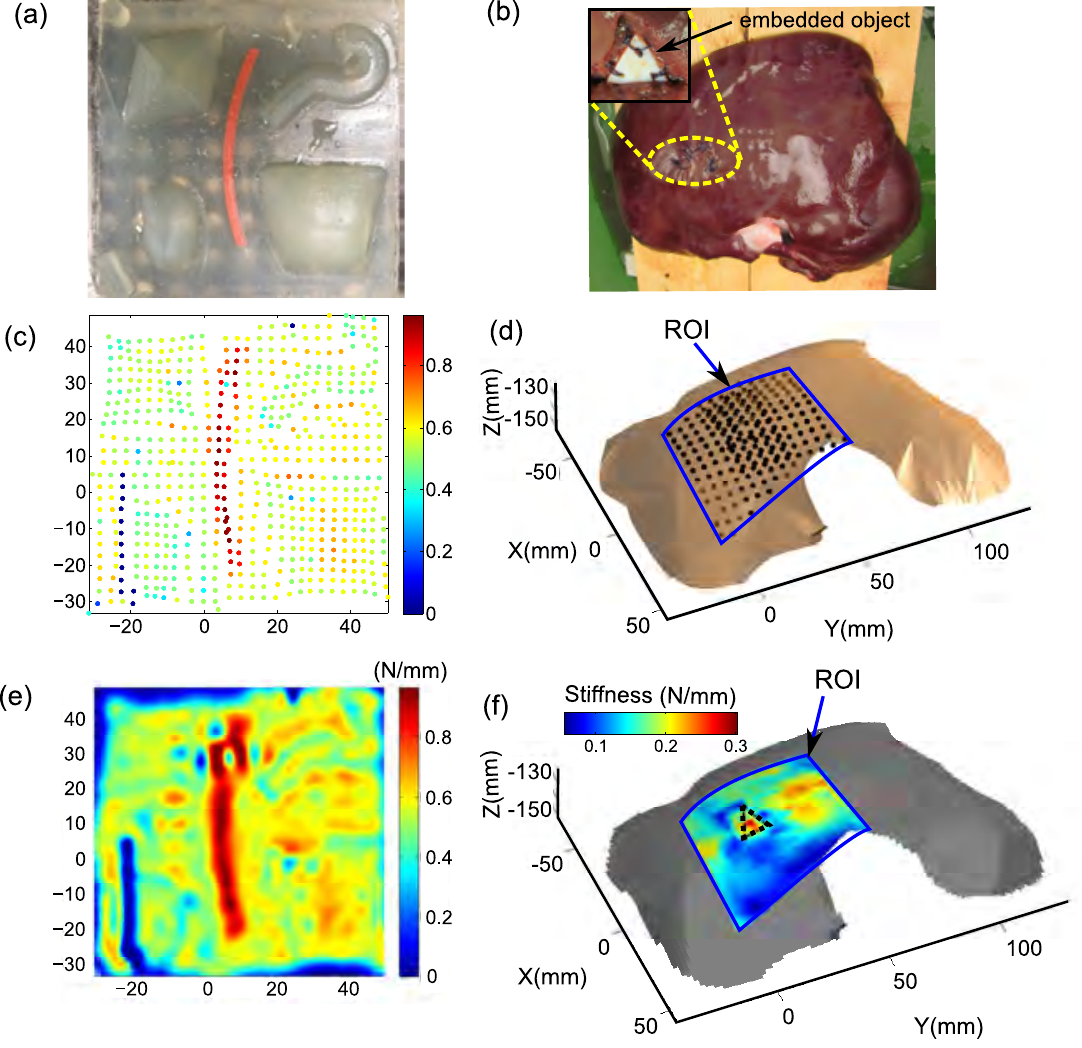} 
      \caption{(a)~Silicon phantom organ (b)~\textit{ex~vivo} porcine liver with an inclusion sutured inside the organ, (c)~619 points were probed on the organ to generate a ground truth for the silicon organ, (d)~196 points were probed on the \textit{ex~vivo} organ to generate a ground truth stiffness map, (e)~Stiffness map used as the ground truth for the silicon organ, (f)~Stiffness map used as the ground truth for the \textit{ex~vivo} organ}
	\label{fig:ExpData}
   \end{figure}

\begin{figure}[tbhp]
      \centering
      \includegraphics[width=\columnwidth]{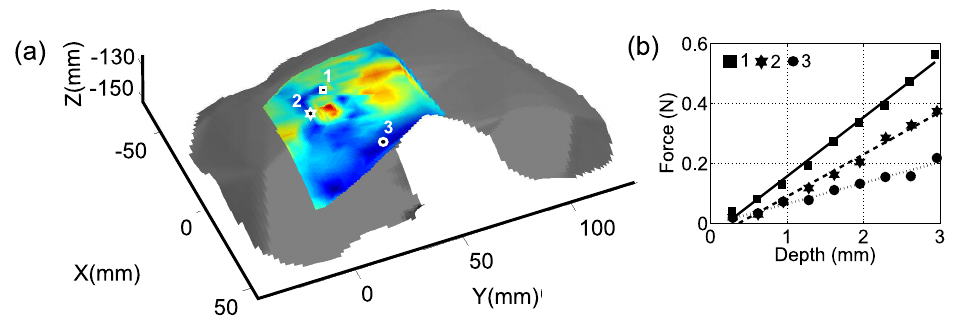} 
      \caption{(a)~CAD model of the organ showing three locations with a square, star and a circle, (b)~Force vs probing depth for the three locations shown on the CAD model.}
	\label{fig:ExpDataStiffness}
   \end{figure}

The silicon organ used in Example 3 and the \textit{ex~vivo} organ used in Example 4 are shown in Fig.~\ref{fig:ExpData}(a) and (b), correspondingly. The ground truth stiffness distribution for Example~3 and for the ROI in Example~4  were obtained by interpolating the experimental data at the grid locations shown in Fig.~\ref{fig:ExpData}(c) and (d) and are shown in Fig.~\ref{fig:ExpData}(e) and (f), respectively. We emphasize that the organ was discretized only to palpate each grid point with the robot for the purpose of generating a ground truth stiffness map of the organ to test our algorithm. Fig.~\ref{fig:ExpDataStiffness}(b) shows that sensed force is proportional to the depth of probing for the three locations shown on the CAD model of the liver in Fig.~\ref{fig:ExpDataStiffness}(a), validating that the linear stiffness is a valid assumption for 3mm probing depth. 
	
\begin{figure}[htbp]
      \centering
      \includegraphics[width=\columnwidth]{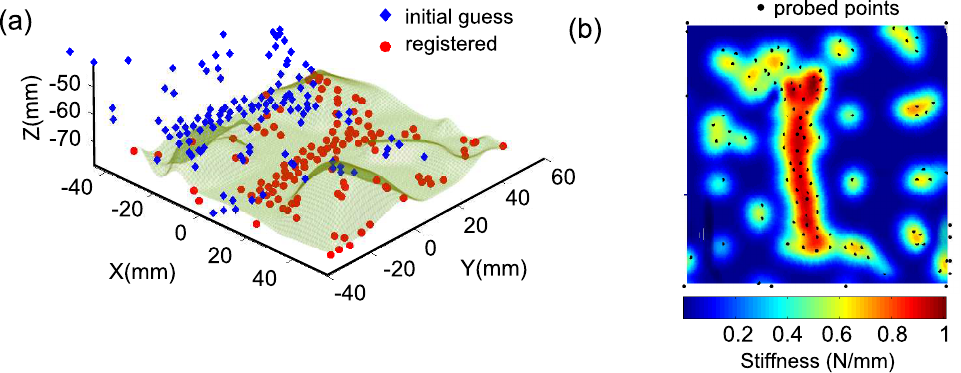} 
      \caption{(a)~Registration results for Example~3, (b)~Estimated stiffness map for Example~3 with 119 probed points. The algorithm starts with an initial grid of 19 points: 4 corners and 15 uniformly spaced points.}
	\label{fig:ExpDataGPRsilicon}
   \end{figure}

\begin{figure}[tbhp]
      \centering
      \includegraphics[width=\columnwidth]{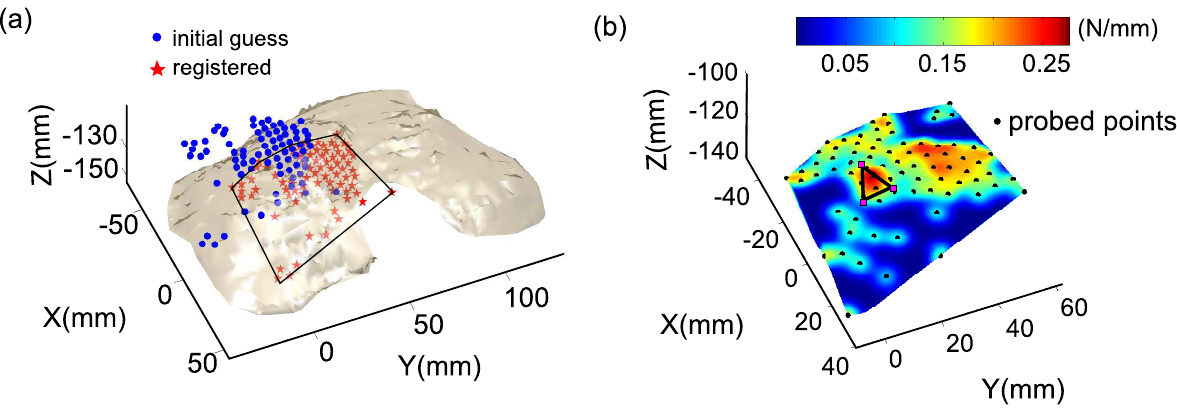} 
      \caption{(a)~Registration results for Example~4, (b)~Estimated stiffness map for Example~4 with 79 probed points. The initial set for Example~4 consists of 7 samples that define the boundary of ROI and 12 uniformly distributed samples inside the ROI.}
	\label{fig:ExpDataGPRexvivo}
   \end{figure}

Fig.~\ref{fig:ExpDataGPRsilicon} shows the successful registration and the estimated stiffness map that reveals the mock artery in the silicon organ. Fig.~\ref{fig:ExpDataGPRexvivo} shows the registration result and the position of the embedded triangle overlayed on the estimated stiffness map of the \textit{ex~vivo} porcine liver.  The actual registration parameters, the estimated registration parameters and the root mean square (RMS) error between the estimated location of all the probed points and their true positions  are shown in Table~\ref{tb:GPR} for all Examples.
\setlength{\tabcolsep}{0.6mm}
\begin{table}[htbp]
\caption{Registration Results}
\label{tb:GPR}
\centering
\begin{tabular}{*2l*7c}
\toprule
{}&Example&x(mm)&y(mm) & z(mm) &$\theta_x$(deg) &$\theta_y$(deg)&$\theta_z$(deg) &RMS(mm)  \\ \midrule
Actual&1-3 & 5&10  &-15  &11.46  &-11.46&5.73  & --   \\
Guess&1-3 &0 & 0 & 0    & 0 &0& 0& -- \\
CMU&1   & 4.51&9.56  &-15.03  &12.57 &-11.42&5.53& 0.91  \\ 
CMU&2   & 4.31&9.55  &-14.97  &12.53 &-11.42&5.53& 1.04  \\ 
CMU&3   & 4.56&9.24  &-14.97  &12.58 &-11.27&5.69 &1.14  \\  \midrule
Actual&4 & 5&7  &-13  &11.45  &-5.72&8.59  & --   \\
Guess&4  &0 & 0 & 0    & 0 &0& 0& -- \\ 
CMU&4   & 5.78&6.4  &-13.04  &11.84 &-5.50 &8.66 &0.74 \\ \bottomrule
\end{tabular}
\end{table}

\section{CONCLUSIONS}
\label{sec:Conclusion}
This work introduced a probabilistic estimation of the stiffness distribution of the organ using Gaussian processes and Bayesian optimization to direct the probing for maximum information gain. We believe fusing intraoperative data into the preoporative model is important to alleviate the limited situational awareness in MIS. The performance of the method was demonstrated by a number of examples and the results show that information-guided probing can avoid probing the entire organ and successfully reveal the stiff regions while registering the tool to the \textit{a priori} geometric model of the organ. 

There are several directions for future work. We used a simple experimental setup and an unconstrained environment to evaluate our method to avoid additional sources of error such as workspace limitation and deflection of the robot. In our future work, we will demonstrate the proposed method in real-time using a continuum robot~\cite{tully2012} and the da Vinci Research Kit. Another extension of this work is the intraoperative reconstruction of the organ surface and stiffness features as the organ goes through changes during the surgical procedure. We envision that the information-guided probing will enable generation of an updated model of the visible anatomy and reduce the time it takes to reconstruct the intraoperative scene.




\section*{APPENDIX}
\subsection{Complementary Model Update for CARE}
\label{appendix}
The CMU method is briefly described here for completeness. The various steps involved in the CMU can be described as follows:
\begin{enumerate}
\item \textit{Collection}:  In the collection step, pairs of force-position measurements which satisfy the following conditions are grouped together in the same set:
\begin{enumerate}[\itshape i)]
\item The force magnitudes are different.
\item The position measurements are spatially close by.
\item The normal directions are the same or similar up to a threshold.
\end{enumerate}
In the experiments described in Section~\ref{sec:Experiments}, we assume that the surface is smooth and frictionless, thus the applied force is along the surface normal and increases with depth. The three conditions stated above imply that distinct measurements that lie on the surface normal experience different force, and form a compatible set. 
The magnitude of the sensed normal forces are denoted by $F_k\in R$, and $\bs{p}^R_k \in R^3$ are the coordinates of the sensed points, where $[\ ]^R$ denotes that the entity is in the tool's reference frame. Given the measurements $(\bs{p}^R_k,F_k),$ $k=1,2,...,l$ obtained so far, we collect compatible sets, $\{{\bs{p}^R_j,F_j\}}_i,  i=1,2,...,n$,  where $n$ is the total number of distinct sets obtained and $j \in N_i$ where $N_i$ is the set that contains the indices of measurements belonging to the $i^{th}$ set.
\item \textit{Stiffness estimation}: For each set $i$ that has at least one pair of force-position measurements, we estimate the local stiffness $c_i$, assuming a linear stiffness model.
\begin{equation}
\label{eq:linearstiffness}
c_i=L\left(\norm{(\bs{p}^R_{\beta})_i-(\bs{p}^R_{\gamma})_i},((F_{\beta})_i-(F_{\gamma})_i)\right)
\end{equation}
where $(\beta,\gamma) \in N_i$, $\beta \neq \gamma$. In Eq.~\ref{eq:linearstiffness}, $L$\textit{(depth,force)} is the function that 
returns the slope of the best line that fits the variation of force with deformation depth. 
\item \textit{Correspondence}: The sensed points $(\bs{p}_\beta^R)_i$ are transformed to the CAD frame using the best registration estimate, \bs{T}. Then, we find $\left(\bs{p}_{i}^C,\bs{n}_i^C\right)=M\left(\bs{T}(\bs{p}^R_{\beta})_i,\bs{\phi}\right)$, where
	$b^C =M\left(\bs{a}^C,\bs{\phi}\right)$ is the rule that finds the closest point $\bs{b}^C \in \bs{\phi}$ to $\bs{a}^C$ and the corresponding normal $n^C$, 
	where $[\ ]^C$ denotes entities represented in the preoperative CAD model's reference frame. 
\item \textit{Minimization}: The following objective function is minimized using Arun's method~\cite{Arun87}
\begin{equation}
\label{eq:objective}
\bs{T}=\argmin\limits_{\bs{T}} \sum \limits_{i=1}^n \norm{\bs{p}_{i}^C-\frac{\bs{n}_i^C (F_\beta)_i}{c_i}-\bs{T}(\bs{p}^R_{\beta})_i}.
\end{equation} 
\item Upon obtaining $\bs{T}$, we loop between the Correspondence and Minimization step until convergence or up to a fixed number of iterations.  
\end{enumerate}
The minimization step can return a local minima when Arun's method~\cite{Arun87} or update step of a Kalman filter~\cite{moghari07} is used. Therefore, we seed the algorithm with multiple initial guesses to overcome the problem of local minima. The detailed derivation of the algorithm is described in ~\cite{ArunICRA2016}.


\bibliography{References}
\bibliographystyle{IEEEtran}



\end{document}